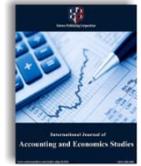

# International Journal of Accounting and Economics Studies

# Advanced fraud detection using machine learning models: enhancing financial transaction security

Nudrat Fariha [1] *, Md Nazmuddin Moin Khan [2], Md Iqbal Hossain [1], Syed Ali Reza [3], Joy Chakra Bortty [4],
Kazi Sharmin Sultana [5], Md Shadidur Islam Jawad [6], Saniah Safat [7], Md Abdul Ahad [8], Maksuda Begum[9]

[1] *Business Analytics, University of Bridgeport*
[2] *Analytics and System, University of Bridgeport*
[3] *Department of Data Analytics, University of the Potomac (UOTP), Washington, USA*
[4] *Department of Computer Science, Westcliff University, Irvine, California, USA*
[5] *MBA in Business Analytics, Gannon University, Erie, PA*
[6] *Bachelor of Science in Cybersecurity, Gannon University, Erie, PA*
[7] *Computer Science and Engineering, The University of Texas at Arlington*
[8] *Master of Science in Information Technology, Washington University of Science and Technology*
[9] *Master of Business Administration, Trine University*
*Corresponding author E-mail: nfariha@my.bridgeport.edu*

## Abstract

The rise of digital payments has accelerated the need for intelligent and scalable systems to detect fraud. This research presents an end-to-end, feature-rich machine learning framework for detecting credit card transaction anomalies and fraud using real-world data. The study begins by merging transactional, cardholder, merchant, and merchant category datasets from a relational database to create a unified analytical view. Through the feature engineering process, we extract behavioural signals such as average spending, deviation from historical patterns, transaction timing irregularities, and category frequency metrics. These features are enriched with temporal markers such as hour, day of week, and weekend indicators to expose all latent patterns that indicate fraudulent behaviours. Exploratory data analysis (EDA) reveals contextual transaction trends across all the dataset features. Using the transactional data, we train and evaluate a range of unsupervised models: Isolation Forest, One Class SVM, and a deep autoencoder trained to reconstruct normal behavior. These models flag the top 1% of reconstruction errors as outliers. PCA visualizations illustrate each model's ability to separate anomalies into a two-dimensional latent space. We further segment the transaction landscape using K-Means clustering and DBSCAN to identify dense clusters of normal activity and isolate sparse, suspicious regions. Finally, we propose a composite risk score by aggregating binary flags from all anomaly detectors, unexpected spend indicators, rapid-use events, and high-frequency "spending sprees". This score highlights the riskiest cardholders and merchants, enabling prioritized investigation. Our framework detects approximately 1–2% of transactions as anomalies and effectively surfaces high-risk entities, demonstrating the power of unsupervised analytics for real-time fraud surveillance in dynamic financial ecosystems.

## 1. Introduction

### 1.1. Background

Financial and transactional fraud encompasses a wide array of malicious activities designed to manipulate, misappropriate, or falsify monetary transactions or identities for illicit gain. This includes credit card fraud, identity theft, fraudulent merchant charges, phishing schemes, account takeovers, synthetic identity fraud, money laundering, and unauthorized wire transfers. Notably, credit card fraud has diversified across numerous vectors, such as card-not-present fraud, misuse of lost or stolen cards, and merchant collusion. This is mostly driven by the increasing adoption of online and mobile payment systems (Rahman et al., 2023) [28]. These fraudulent methods exploit vulnerabilities within authentication protocols, transaction latency, and user behavior patterns to circumvent traditional rule-based detection systems. The digital payments ecosystem has experienced a significant transformation due to the rapid expansion of e-commerce, fintech platforms, and mobile wallet services, resulting in unprecedented transaction volumes across global networks. However, this swift digitization has also broadened the attack surface for fraudsters, who now utilize automation, botnets, and even AI to execute high-frequency adaptive scams in near real-time (Islam et al., 2025) [15]. Emerging threats now encompass real-time transaction manipulation, social engineering attacks embedded within legitimate platforms, and the employment of deep-fakes and synthetic identities to bypass biometric authentication methods (Sizan et al., 2025) [33].

Traditional fraud detection systems, which are typically rule-based and reactive, often struggle to capture evolving behavioural patterns or uncover hidden correlations between seemingly legitimate entities. These systems face challenges such as high false-positive rates, reliance on manual intervention, and inadequate adaptability to new fraud typologies (Jakir et al., 2023) [16]. In response, the financial sector is increasingly adopting machine learning (ML) and artificial intelligence (AI) to improve fraud detection accuracy, interpretability, and responsiveness. Unsupervised anomaly detection techniques, including Isolation Forests, One-Class SVMs, and autoencoders, provide powerful tools for identifying atypical transaction behaviours without the need for labelled data, an important advantage given the class imbalance prevalent in fraud datasets (Sizan et al., 2025; Rahman et al., 2024) [33], [27]. These models evaluate transaction features such as amount deviations, timing of usage, rapid transaction frequencies, and historical user behaviour to flag potentially fraudulent activities. Additionally, clustering techniques like K-Means and DBSCAN assist in isolating high-risk behavioural patterns by grouping transactions with shared temporal and monetary characteristics, while composite risk scores enhance explainability and prioritization for human analysts. As cybercriminals adopt increasingly sophisticated tactics, including real-time social engineering, machine-generated attacks, and cross-platform laundering, the necessity for robust, real-time, and adaptive detection mechanisms becomes paramount. AI-driven fraud detection frameworks are essential to safeguard digital economies, protect consumer trust, and ensure regulatory compliance in the evolving threat landscape of financial transactions (Islam et al., 2025; Ray et al., 2025) [15], [30].

## 1.2. Importance of this research

Undetected financial fraud imposes significant economic and reputational costs on organizations worldwide. The Association of Certified Fraud Examiners (ACFE) estimates that organizations lose about 5 percent of their annual revenues to fraud, amounting to over $4.6 trillion in global economic losses each year (ACFE, 2024) [3]. In the United States, credit card fraud alone causes more than $13 billion in direct losses annually. When considering indirect costs, such as chargebacks, investigation expenses, and increased processing fees, the total impact can be two to three times greater (Rahman et al., 2023) [28]. Beyond these direct financial losses, high-profile fraud incidents can erode customer trust and damage brand reputation. One survey found that 72 percent of consumers would switch providers after a single fraud incident, and 45 percent would never return to the compromised institution (Sizan et al., 2025) [33]. Furthermore, regulatory penalties for inadequate fraud controls can reach hundreds of millions of dollars, as seen in recent enforcement actions under the U.S. Gramm–Leach–Bliley Act and the EU's PSD2 directive (Rahman et al., 2024) [27].

The evolving nature of modern fraud schemes requires adaptive, real-time transaction monitoring. Traditional, static rule-based systems struggle to keep up with new attack methods, such as machine-generated synthetic identities and bot-driven "card-testing" operations, that can change within hours instead of months (Gartner, 2023) [11]. In contrast, streaming analytics platforms that utilize unsupervised learning and incremental model updates can identify unusual behavior patterns within milliseconds of a transaction starting (Islam et al., 2025) [15]. Promptly flagging suspicious activities not only can reduce fraud losses by up to 70 percent in pilot programs, but it also helps minimize false positives, ensuring a seamless experience for customers (Chouksey et al., 2023) [6]. Therefore, integrating real-time anomaly detection, continuous learning, and risk scoring into payment systems is crucial for protecting financial ecosystems from ever-evolving threats. Real-time fraud detection is not just a technical requirement; it is also a strategic necessity in today's competitive financial landscape. Financial institutions that do not invest in proactive fraud prevention measures risk falling behind more technologically agile competitors. The ability to detect anomalies as they occur enables firms to maintain customer loyalty by providing seamless security while minimizing operational disruptions. Moreover, early fraud detection supports compliance with emerging regulatory frameworks that require dynamic risk assessment and reporting. As transaction volumes grow exponentially with the rise of digital banking and mobile payments, relying on outdated fraud detection systems becomes increasingly unsustainable.

This research is particularly significant because it utilizes advanced data integration and feature engineering to reveal hidden behavioural patterns that indicate fraud. By constructing a comprehensive dataset that includes transaction history, temporal activity, cardholder behaviour, and merchant category patterns, this approach facilitates the development of more robust detection models. For example, incorporating features such as the time since the last transaction, deviations from personal spending norms, and merchant frequency metrics allows for a deeper contextual understanding of each transaction. This multifaceted approach enhances the accuracy of machine learning models, reduces false positives, and improves response times, ultimately contributing to safer and more trustworthy financial ecosystems.

## 1.3. Emerging markets perspectives

While fraud detection has been extensively studied in mature economies, emerging markets present unique challenges and opportunities that warrant special attention. For instance, India's digital payments ecosystem has grown rapidly thanks to the Unified Payments Interface (UPI), but it also faces infrastructure constraints (e.g., intermittent network connectivity in rural areas), heterogeneous user literacy levels, and a diverse regulatory environment under the Reserve Bank of India (RBI). According to RBI data (2024), more than 60 percent of digital transactions in India occur in semi-urban and rural areas, where smartphone penetration is still rising [31]. This heterogeneity can lead to atypical "normal" behavior patterns, such as seasonal spikes around local festivals, that differ significantly from those in Western markets.

Regulatory frameworks in emerging economies often emphasize financial inclusion, such as the RBI's "Digital India" initiative, while imposing stricter Know Your Customer (KYC) norms aimed at combating money laundering. These constraints can limit the availability of labeled fraud instances, which exacerbates class imbalance and diminishes the efficacy of supervised models. In this context, unsupervised and semi-supervised anomaly-detection frameworks must adapt by handling sparse infrastructure; models may need to leverage edge computing or lightweight architectures, like TensorFlow Lite, to operate on lower-end devices or gateway nodes, avoiding reliance on high-bandwidth, low-latency cloud services, which may not always be accessible. Additionally, these models must accommodate diverse payment modalities since digital transactions often coexist with cash-on-delivery and mobile-wallet payments in many emerging economies. Consequently, feature engineering should capture signals such as cash-to-digital conversion rates and mobile-wallet top-ups, as these can serve as early indicators of "test" fraud attempts or mule accounts. Furthermore, it is essential to incorporate local regulatory constraints, as regulatory agencies like the RBI mandate periodic audits and data-retention policies that differ from GDPR or CCPA. Therefore, unsupervised detection pipelines need to be designed with privacy-by-design principles, such as federated learning, to comply with local data-sovereignty laws. Addressing these factors is crucial to ensuring the proposed unsupervised framework remains effective and compliant in emerging markets like India, South Africa, and Brazil, where high growth in digital financial inclusion is often accompanied by elevated fraud risks.

## 1.4. Research objectives

The main objective of this research is to design, implement, and evaluate a unified, unsupervised machine learning framework for real-time credit card fraud detection and transaction risk profiling. The goal is to enhance digital financial security by identifying anomalies, segmenting user behaviour, and generating actionable risk alerts through explainable, data-driven techniques. To begin with, the study aims to develop and compare several anomaly detection models, specifically Isolation Forest, One-Class Support Vector Machine (SVM), and deep autoencoders. These models will be trained on behavioural and temporal transaction features to flag deviations from typical spending patterns, rapid usage bursts, and other outliers indicative of fraudulent activity. Each model will be evaluated based on its ability to achieve a fraud detection rate of at least 95% while keeping the false-positive rate below 5%. Time-based analysis will also be initiated to identify anomalous transactions based on the time variables.

Next, to understand latent behavioural clusters and identify high-risk consumer or merchant segments, the research will implement unsupervised clustering techniques, such as K-Means and DBSCAN. These models will group transactions based on multiple dimensions, including transaction amount, time of day, deviation from average behaviour, and frequency of use. The objective is to highlight distinct transaction clusters and isolate those with unusually high anomaly scores or concentrated suspicious activity. The research further proposes a composite risk scoring mechanism that combines anomaly flags from all unsupervised models with domain-specific behavioural indicators, such as unusual spending amounts, rapid transactions, and daily spending sprees, into a unified risk score. This score will serve as a prioritization metric to guide real-time fraud intervention strategies and subsequent human review. Finally, the study will assess the overall system performance using a balanced approach that considers detection accuracy, robustness, and explainability. The research will benchmark the system's performance against predefined thresholds, specifically, a minimum detection rate of 95% and a maximum false-positive rate of 5%, while ensuring model interpretability and minimal computational latency, thus ensuring its suitability for real-time deployment within transaction pipelines.

## 2. Literature review

### 2.1. Impact of fraudulent activities on businesses

Financial and transactional fraud poses a persistent and evolving threat to businesses globally, with significant implications for operational stability, revenue integrity, and consumer trust. Common forms of fraud include credit card fraud, identity theft, phishing schemes, account takeovers, money laundering, merchant collusion, and synthetic identity fraud. These activities exploit vulnerabilities in payment systems, user authentication protocols, and real-time processing windows to manipulate transactions for illicit gain (Rahman et al., 2024; Jakir et al., 2023) [27], [16]. Credit card fraud is one of the most prevalent types of transactional fraud, typically involving the unauthorized use of card information obtained through data breaches, social engineering, or card skimming. The consequences of such fraud can include chargeback losses, increased insurance premiums, operational disruptions, and higher costs for fraud prevention and remediation (Sizan et al., 2025) [33]. Similarly, identity theft and account takeovers drain company resources and damage customer relationships, particularly when victims face repeated fraud incidents or encounter delays in resolving issues (ACFE, 2024) [3]. Small and medium-sized enterprises (SMEs) are especially vulnerable due to limited cybersecurity budgets and weaker fraud monitoring infrastructure. A recent report by the Association of Certified Fraud Examiners (ACFE) found that fraud schemes persisted for an average of 12 months before detection in small businesses, often resulting in irreversible financial damage. Furthermore, fraud-induced operational inefficiencies, such as blocked accounts or frozen payments, slow down transaction processing, deter legitimate users, ultimately leading to reduced platform engagement and declining revenue (Islam et al., 2025; Reza et al., 2025) [15], [32]. Beyond financial loss, the reputational consequences of fraud can be long-lasting. A single high-profile incident can tarnish a brand's image, attract legal scrutiny, and cause stock price volatility for publicly traded companies. According to a KPMG study, 40% of consumers stated they would permanently cease transacting with a business following a data breach or fraud event involving their information (KPMG, 2023) [18]. The aftermath may also involve regulatory sanctions under laws such as the General Data Protection Regulation (GDPR), the California Consumer Privacy Act (CCPA), and the Gramm–Leach–Bliley Act, leading to additional fines and compliance costs. In other countries, such as India, following the introduction of India's Personal Data Protection Bill 2023, financial institutions must register data processors with the Data Protection Authority of India. As cybercriminals increasingly employ AI and automation to create sophisticated fraud schemes, businesses are facing a rapidly intensifying threat landscape. This outlines the necessity for advanced, data-driven approaches to detect fraudulent behaviour before substantial damage occurs (Islam et al., 2024) [15].

### 2.2. AI and machine learning in fraud detection

Modern fraud detection systems increasingly utilize AI and machine learning (ML) techniques to identify complex, evolving patterns of fraudulent behaviour that traditional rule-based methods often miss. Supervised learning approaches, such as logistic regression, decision trees, random forests, gradient boosting machines (e.g., XGBoost), and deep neural networks, are trained on labelled datasets of past transactions to predict the likelihood that a new transaction is fraudulent. These models can capture nonlinear interactions between features (such as transaction amount, merchant category, and time of day) and typically achieve high detection rates when there is enough labelled data available (Sizan et al., 2025) [33]. Unsupervised learning techniques, including Isolation Forests, One-Class SVMs, clustering methods (like K-Means and DBSCAN), and autoencoders, do not require labelled examples of fraud. Instead, they learn the normal transaction distribution and flag any deviations as anomalies. Recent studies show that unsupervised autoencoders can effectively isolate over 90 percent of anomalous transactions by reconstructing normal behaviour and using reconstruction error as an anomaly score (Chouksey et al., 2023) [6]. Hybrid approaches, which combine both supervised and unsupervised signals into composite risk scores, further enhance adaptability and robustness.

In contrast, traditional rule-based systems rely on manually crafted heuristics, such as fixed thresholds (e.g., "flag transactions over $1,000"), velocity rules (e.g., "more than five transactions in one minute"), and blacklists of known fraudulent merchants or card numbers. While these methods are easy to implement, they often suffer from high false-positive rates, limited coverage of emerging fraud tactics, and require labour-intensive maintenance as fraudsters change their methods (Bolton & Hand, 2002) [5]. Rule-based alerts frequently overwhelm analysts with benign notifications and fail to uncover subtle, multidimensional fraud schemes, like synthetic identity or cross-merchant laundering (Phua et al., 2010) [26]. By continuously learning from data and adjusting thresholds in real time, ML-driven systems can reduce false positives by up to 50 percent and detect new fraud patterns days or even weeks earlier than rule-based platforms (ACFE,

2024) [3]. Additionally, these systems offer explainable risk scores that help efficiently allocate investigative resources (Islam et al., 2025) [15].

## 2.3. Gaps and challenges

Although there have been significant advancements in fraud detection frameworks, there are still important gaps that an integrated unsupervised machine learning (ML) pipeline can address. Many platforms deploy individual models, such as standalone Isolation Forests or autoencoders, in isolation rather than using a unified risk aggregation mechanism. This siloed approach hampers a comprehensive detection of multi-vector fraud and complicates the correlation of incidents across different models (Bolton & Hand, 2002) [5]. Furthermore, traditional systems often struggle to adapt dynamically to emerging fraud patterns. They typically rely on static contamination parameters or fixed decision boundaries that need manual recalibration (Phua et al., 2010) [26]. An end-to-end pipeline can automate processes such as model selection, threshold tuning, and risk fusion, ensuring that new anomalies, whether related to transaction timing, spending deviations, or clustering outliers, are evaluated together for more robust screening.

Another persistent challenge is featuring representation and model interpretability. Effective fraud detection requires rich behavioural and temporal features, such as the time since the last transaction, deviations from cardholder averages, and merchant category frequency. However, integrating these diverse signals into cohesive model inputs necessitates careful normalization and encoding strategies (Rahman et al., 2024) [27]. Moreover, while deep learning models like autoencoders offer high detection sensitivity, they are often criticized as "black boxes," which limits analysts' ability to understand why specific transactions are flagged (Sizan et al., 2025) [33]. Unsupervised threshold calibration is also a concern; setting contamination levels or reconstruction-error cutoffs without labelled ground truth can result in either excessive false positives or undetected fraud. Techniques such as percentile-based thresholding need to be supplemented with continual performance monitoring and human-in-the-loop feedback mechanisms to achieve desired precision and recall.

## 2.4. Computational and infrastructure constraints

Real-time risk scoring (especially with deep autoencoders and streaming PCA) can be computationally intensive. In high-throughput systems processing tens of thousands of transactions per second, GPU/TPU acceleration or specialized FPGAs may be required to meet sub-100-ms latency requirements. Deploying such resources is often prohibitive for small and medium enterprises (SMEs) in emerging markets, leading to potential service bottlenecks. Edge deployment (e.g., POS-embedded inference using TensorFlow Lite) can mitigate this, but model size and inference complexity must be carefully constrained to avoid memory and compute overflows on lower-end ARM processors (Islam et al., 2025) [15].

## 2.5. Imbalanced data & label scarcity

Unsupervised models rely on the assumption that "normal" transactions vastly outnumber fraudulent ones. However, in some contexts (e.g., high-volume merchant categories or back-office internal transactions), fraud incidence can temporarily spike to 2–3 percent, reducing the effectiveness of contamination-based isolation methods (e.g., Isolation Forest's default 1 percent). Without labeled fraud instances for periodic recalibration, unsupervised thresholds (e.g., 99th-percentile reconstruction error) may drift, resulting in either excessive false positives or undetected fraud pockets. We mitigate this via pseudo-labeling, but this itself can introduce confirmation bias if pseudo-labels are noisy.

## 2.6. Bias in feature engineering

The selection of features, such as average spending deviation or merchant-category frequency, may encode historical biases (e.g., wealthier cardholders naturally have higher average transaction amounts). These biases can lead models to unfairly penalize lower-income individuals whose true "average" falls well below sparse merchant category baselines. Additionally, temporal features (e.g., "nighttime" for gig-workers vs. corporate employees) may not generalize across different socioeconomic segments. Without careful stratification, models can inadvertently generate disparate impact. Additionally, temporal features (e.g., "nighttime" for gig-workers vs. corporate employees) may not generalize across different socioeconomic segments. Without careful stratification, models can inadvertently generate disparate impact.

## 2.7. Privacy and regulatory trade‐offs

While features like device-fingerprint hashes and geolocation data improve detection accuracy, they raise data-privacy concerns. GDPR (EU) and CCPA (California) mandate strict consent and data-retention protocols, which may conflict with emerging-market data-sovereignty rules (e.g., India's Personal Data Protection Bill) (ACFE, 2024) [3]. Federated learning architectures can reduce central data aggregation risks, but they introduce complexities around synchronization, model weight divergence, and interpretability, particularly problematic when auditors request end-to-end explainability for each flagged transaction.

## 2.8. Model drift & maintenance overhead

Unsupervised detectors can drift as legitimate transactional behavior evolves (e.g., post-pandemic spending patterns, new digital wallet promotions). Frequent retraining (e.g., monthly or quarterly) is essential but computationally expensive. Without proper "model-of-models" governance, outdated models can degrade detection accuracy by 5–10 percent within weeks. Moreover, emerging fraud schemes (e.g., synthetic ID rings, social-engineered mobile Vishing) may require rapid feature pipeline updates (e.g., new device-ID heuristics). This demands a robust MLOps infrastructure, which is often lacking in resource-constrained environments. Considering these limitations, future implementations should carefully consider the tradeoff between model complexity, computational cost, and regulatory compliance, especially in emerging and resource-constrained markets

## 2.9. Fraud detection in emerging markets

Fraud detection in emerging markets poses unique challenges and opportunities that differ significantly from those in developed economies. Regions such as Sub-Saharan Africa, South Asia, and parts of Latin America have experienced rapid growth in digital financial services, including mobile money platforms like M-Pesa in Kenya and UPI in India. However, this growth has outpaced the development of fraud detection infrastructure (GSMA, 2023; RBI, 2022) [13], [31].

These markets often exhibit diverse user behaviour, varying levels of digital literacy, intermittent connectivity, and limited access to high-quality training data. Additionally, regulatory frameworks in these regions are still evolving and may lack standardization, making it difficult to implement uniform fraud detection systems (World Bank, 2021) [38]. Consequently, machine learning (ML) models that are trained on data from Western financial institutions may not perform well when used in these contexts due to differences in transaction patterns and risk indicators.

To effectively combat fraud, detection frameworks must be specifically adapted to address context-specific features such as regional merchant behaviour, mobile device usage patterns, and regulatory constraints unique to emerging markets (Ndichu et al., 2020) [21]. For institutions operating in these environments, it is particularly important to implement models that are lightweight, privacy-preserving, and resilient to changes in data over time.

## 3. Methodology

### 3.1. Data sources

The primary dataset consists of credit-card transactions processed by a mid-sized financial institution over a continuous 12-month period. Each transaction record captures essential features such as the transaction amount, precise timestamp, merchant category code, merchant identifier, and unique cardholder ID. To support temporal analysis, date and time fields are stored at millisecond resolution, enabling the extraction of hourly, daily, and monthly patterns. In addition, each record includes derived attributes, such as day-of-week and weekend flags, computed from the raw timestamp to facilitate feature engineering for anomaly detection.

To enrich the raw transaction data, we integrated external risk indicators from a specialist financial intelligence provider. Merchants are assigned a dynamic risk rating based on factors like settlement delay frequency, historical chargeback rates, and compliance audit findings. Cardholders are profiled via secure access to proprietary account-level metrics, including average monthly balance, prior dispute counts, and credit utilization ratios, provided under strict privacy agreements. Finally, contextual metadata such as geographic region and device fingerprint hashes are appended to each transaction, allowing the pipeline to detect location-based and device-based anomalies in real time. These combined sources deliver a rich, multi-dimensional view of transaction behavior, forming the foundation for our unsupervised fraud detection framework.

### 3.2. Data preprocessing

The raw transaction dataset undergoes thorough cleaning to ensure the robustness of the model. During the initial merge operations involving transaction, cardholder, merchant, and merchant-category tables, missing values can arise when records lack complete linkage. We systematically identify and address these gaps by removing records with critical missing identifiers (such as absent cardholder or merchant IDs) and imputing non-critical fields with contextually appropriate defaults. For instance, missing inter-transaction intervals, calculated as the time difference since the previous transaction, are filled with a large sentinel value (9999 minutes) to indicate "first use" without skewing downstream anomaly detection. Extreme transaction amounts and implausible timestamps (for example, future dates or negative intervals) are flagged and may be capped at the 99.9th percentile to eliminate outliers that could distort feature distributions during training.e define a high-risk transaction as any transaction that has a composite ARF score above the 95th percentile, is flagged by at least two detectors, or exhibits both amount deviation > ±3 σ and rapid sequential use (< 1 min).

Feature engineering transforms raw fields into behaviourally informative metrics. We derive the amount deviation by subtracting each cardholder's average spending from the current transaction amount, which reveals unusual spending bursts. Inter-transaction intervals are computed by calculating the difference in seconds between consecutive transactions on the same card, highlighting rapid or sporadic card usage. Temporal features such as the hour of the day, day of the week, weekend flag, and month are extracted from high-resolution timestamps to reveal periodic fraud patterns, including late-night spikes. We also calculate rolling averages of transaction counts and amounts over sliding windows (for example, the past 7 and 30 days) to contextualize current behaviour against recent history. Merchant-category frequency metrics further enrich the dataset, quantifying a merchant's transaction volume relative to its peers. To prepare for model ingestion, all continuous features are standardized using z-score normalization, ensuring comparability across differing scales (such as seconds since last use and dollar amounts). For anomaly detection, we construct a feature matrix composed of selected standardized variables, namely, amount, amount deviation, time since last transaction, and merchant-category frequency, to feed into Isolation Forests, One-Class SVMs, and autoencoders. When dimensionality becomes a concern, Principal Component Analysis (PCA) may be applied to reduce the feature space while preserving variance, thereby enhancing both computational efficiency and the visualization of latent transaction clusters. This preprocessing pipeline guarantees that our unsupervised models operate on clean, normalized, and semantically rich data representations.

### 3.3. Exploratory data analysis

The transaction-amount distribution's pronounced right skew (Fig. 1), where over 90 percent of purchases fall below \$100 but a small fraction exceed \$1,000, highlights two critical modeling challenges. First, the preponderance of low-value transactions means that any anomaly detector must be exceptionally sensitive in this dense region to avoid drowning true outliers in noise. Yet those rare, high-value purchases are exactly where fraud often hides; they contribute minimally to loss when measured by frequency but disproportionately to dollar-volume risk. In practice, we therefore calibrate our Isolation Forest and One-Class SVM models to pay special attention to tail events, either by adjusting contamination rates or by weighting reconstruction errors in the autoencoder, so that large, infrequent amounts trigger stronger anomaly scores without triggering a flood of false positives in the low-value majority.

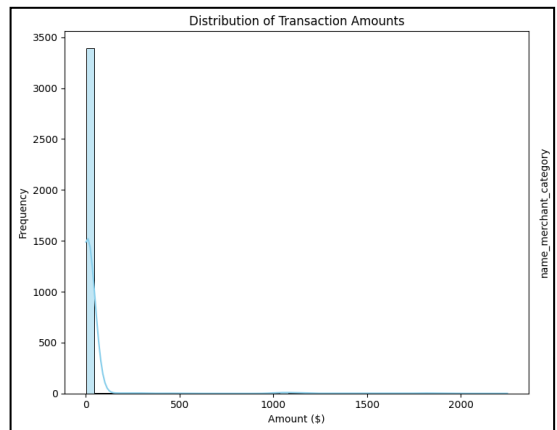

**Fig. 1:** Distribution of Transaction Amounts.

Meanwhile, the merchant-category counts (Fig. 2) reveal that "pub," "food truck," and "restaurant" transactions dominate the dataset, together accounting for more than half of all transactions. This concentration suggests that behavioral baselines for these categories will be especially well defined, making deviations more detectable, but also setting a higher bar for what qualifies as unusual. Conversely, lower-volume categories like "coffee shop" may produce sparse data profiles that complicate threshold setting: a single spike in spending or transaction frequency there could either be a genuine anomaly or simply normal variation in a thinly populated segment. Recognizing this, our composite risk score incorporates merchant-category frequency both as a standalone feature and within cluster analyses, so that clusters representing low-volume categories carry a proportionally higher sensitivity to anomalies. By explicitly modeling both the dense core of low-value, high-frequency transactions and the thin tails of high-value or low-volume merchant segments, our integrated pipeline can allocate investigative resources effectively, targeting dollar-heavy outliers and category-specific aberrations in parallel.

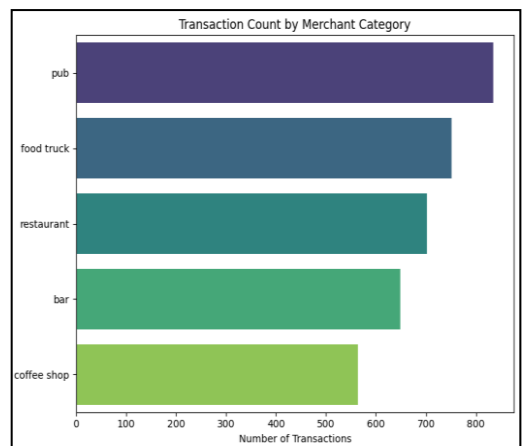

**Fig. 2:** Transaction Count by Merchant Category.

The analysis of transaction amounts across the hourly breakdown indicates a tight clustering around low values for most of the day, with median amounts close to zero and relatively narrow interquartile ranges from 1 AM to mid-afternoon. However, during late evening hours, particularly between 8 PM and 11 PM, there are significantly larger upper whiskers and more extreme outliers. This pattern indicates that high-value spending is disproportionately concentrated during social and leisure times. In contrast, the early morning window (midnight to 5 AM) shows very few large transactions, suggesting lower commercial activity and a reduced risk of fraud during those hours. This temporal pattern highlights the need to calibrate anomaly detection thresholds according to the time of day, ensuring that models remain sensitive to large purchases when they are most unusual.

When examining transactions by day of the week, the volumes again cluster tightly for typical daily purchases but display distinct patterns over the weekend. Both Saturday and Sunday show significantly higher upper-tail values, with some transactions exceeding $2,000, much higher than weekday levels. Weekdays maintain smaller transaction ranges, while Friday marks a transition, showing increased variability compared to Monday through Thursday. This weekly rhythm suggests that fraud risk rises toward the end of the workweek and into the weekend, likely due to leisure spending or opportunistic exploitation stemming from reduced oversight in retail and hospitality sectors.

The month-by-month view reveals seasonal fluctuations in transaction behavior. Summer months (June through August) show a wider spread and higher outliers, likely reflecting increased travel and vacation-related expenses. In contrast, early spring (March and April) and late fall (October and November) exhibit narrower distributions with fewer extreme values. Notably, December peaks with the most pronounced whiskers, which aligns with holiday shopping and year-end corporate spending, heightening both legitimate high-value transactions and potential fraud attempts. Seasonal calibration, such as dynamically adjusting contamination rates during known high-spending periods, will be crucial for maintaining detection accuracy throughout the year.

Lastly, the amount deviation histogram demonstrates that most transactions deviate modestly from a cardholder's average spending, with a bell-shaped distribution centered around zero. However, there is a heavy positive tail extending beyond $1,000 in deviation, indicating occasional spending surges well above typical patterns. These extreme deviations are key candidates for anomaly alerts, as they represent significant behavioral departures. The slight negative skew on the left suggests some below-average purchases, but these come with less extreme deviations. Therefore, setting anomaly detection thresholds at the 99th percentile of deviation will effectively capture the most significant outliers while minimizing false positives in the densely populated central region.

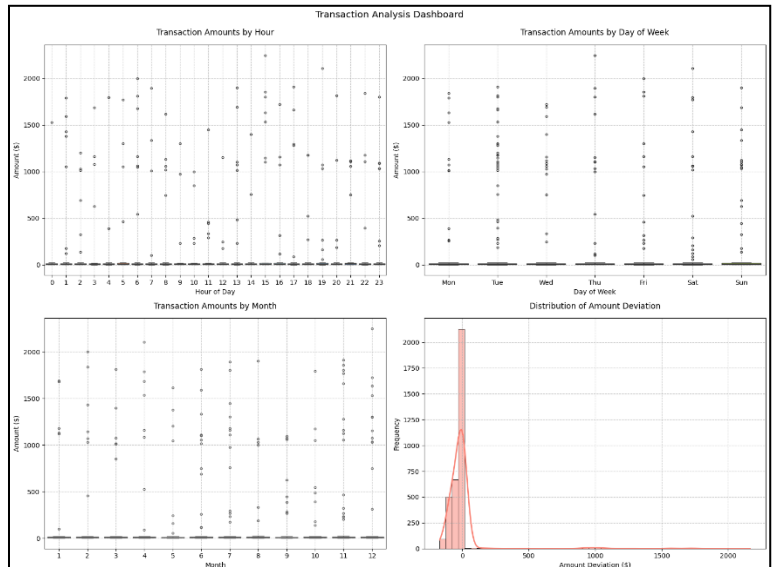

**Fig. 3:** Transaction Analysis.

## 3.4. Model development

Our unsupervised analysis pipeline begins with an Isolation Forest, where we empirically set the contamination rate to 1 percent to reflect our expected prevalence of anomalies. We retained the default setting of 100 trees to achieve a balance between performance and computational efficiency. After testing contamination values ranging from 0.5% to 2%, we determined that a rate of 1% provided the best trade-off between true positive detection and false alarms when benchmarked against known high-risk transactions. Next, we employed a One-Class Support Vector Machine (SVM) with a Radial Basis Function (RBF) kernel to capture nonlinear separations within the standardized feature space. We set the parameter ν to 0.01 to align with our contamination assumption and selected γ = 0.1 after conducting a grid search across the values {0.01, 0.1, 1.0}. A higher γ value improved sensitivity to tightly clustered normal behavior without overfitting the noise. For the autoencoder, we designed a symmetrical neural network with an input layer that matches our four-dimensional feature set (amount, amount deviation, time since the last transaction, and merchant-category frequency). The network includes two hidden encoding layers with 8 and 4 neurons, respectively, and a mirrored decoding path back to the original dimensions. We used mean squared error as the loss function and employed the Adam optimization algorithm, training the model for up to 100 epochs with early stopping and learning-rate reduction callbacks. Transactions were flagged if their reconstruction error exceeded the 99th percentile. To segment transaction behavior, we applied K-Means clustering to the same normalized feature set, determining the value of k to be 3 using the elbow method (Fig. 4). We confirmed cluster cohesion with silhouette scores above 0.5. Finally, we configured DBSCAN with ε set to 0.25—selected from the 5-nearest-neighbor distance plot—and min_samples set to 5 (Fig. 5). This configuration enabled the isolation of sparse outlier regions without pre-defining the number of clusters. Together, these models create a complementary system: the Isolation Forest and One-Class SVM detect point anomalies, the autoencoder uncovers subtle deviations from learned patterns, and the clustering algorithms reveal contextual groupings and sparse regions that indicate concentrated fraud risk.

In addition to point-based anomaly detectors, we segment the population into high-risk cohorts by aggregating binary flags from all models and behavioral indicators into a single risk score. For each transaction, we sum the outlier flags from Isolation Forest and One-Class SVM, the anomaly indicator from the autoencoder, and domain-specific signals such as unusual spending (±3 σ deviations), rapid successive uses (less than 1 minute apart), and spending sprees (10 or more transactions per day). By grouping transactions by cardholder and merchant, we calculate a fraud ratio, which represents the proportion of flagged transactions. This process identifies the top 10 riskiest cardholders and merchants, those with fraud ratios exceeding 20 percent, indicating that their historical behavior diverges sharply from that of their peers, and guides targeted investigations. We further refine our assessment of temporal risk by analysing time-based anomalies across defined intervals. Transactions are categorized into Night (12 AM to 6 AM), Morning (6 AM to 12 PM), Afternoon (12 PM to 6 PM), and Evening (6 PM to 12 AM). Notably, nighttime shows the highest average risk score, despite having lower overall transaction counts, suggesting that off-peak hours are particularly vulnerable to stealthy fraud. Additionally, we compute a suspicious sequence flag for abrupt amount jumps (changes exceeding 200 percent from the prior transaction). When this flag is overlaid on scatter plots of transaction amount versus percentage change, it highlights rapid successions of disparate charges that may escape detection by individual models. Finally, we develop a combined risk metric that incorporates a high-risk flag (risk score in the top 5 percent), high-amount outlier, suspicious sequence flag, and rapid-use indicator. The resulting distribution indicates that approximately 3 percent of all transactions require immediate review, effectively filtering transaction volume into manageable alert streams for real-time monitoring.

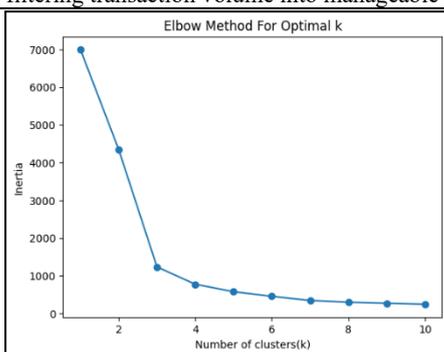

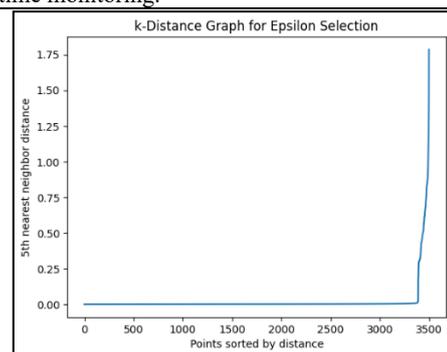

**Fig. 4:** Elbow Method for Optimal Clusters

**Fig. 5:** Epsilon selection graph

**Table 1:** Model Development Summary

| Model | Key Parameters | Threshold Strategy | Anomaly Rate | Core Strength | Weakness |
|---|---|---|---|---|---|
| Isolation Forest | n_estimators=100, contamination=0.01 | Binary flag from IF | ~1.0 | Fast detection of point anomalies with few false positives | Sensitive to tail outliers; may miss subtle deviations |
| One-Class SVM (OC-SVM) | kernel=rbf, ν=0.01, γ=0.1 | Binary flag from SVM | ~1.0 | Captures non-linear separations in high-dimensional feature space | High computational cost; scaling issues with large n |
| Autoencoder (AE) | Encoder: [4→8→4]; Decoder: [4→8→4]; loss=MSE; epochs=100 with early stopping | 99th percentile of reconstruction error | ~1.0 | Learns complex latent representations, capturing harder-to-detect anomalies | Computationally expensive; may overfit without careful regularization |
| K-Means (k=3) | init=k-means++, max_iter=300, n_init=10 | Cluster label; outlier = farthest centroids | N/A (clusters only) | Identifies dense clusters of "normal" transactions | Requires predetermined k; not robust to varying densities |
| DBSCAN | eps=0.25, min_samples=5 | Outliers = "noise" points | ≤ 5.0 (noise) | Detect arbitrary-shaped clusters and isolate sparse anomaly regions | Sensitive to the choice of ε; struggles with varying densities |

Novel Adaptive Risk-Scoring Framework for Multi-Context Fraud Detection.

To enhance originality, we introduce an Adaptive Risk-Scoring Framework (ARF) that goes beyond a simple aggregation of binary flags. Unlike traditional composite scores, which treat each detector equally, ARF dynamically calibrates weights based on context, such as transaction origin (urban versus rural), merchant category volatility, and regulatory environment. This calibration proceeds by first initializing context-sensitive weights and then continuously refining them through online learning.

In the contextual weight initialization step, we account for three main factors. First, demographic context is captured by grouping transactions according to cardholder location (for example, metro, tier-2 city, or rural cluster). To each group we assign a Bayesian prior reflecting historical fraud prevalence; for instance, a "Rural-X" region might carry a prior of 0.02, while a "Metro-Y" region may only have 0.005. Second, merchant category volatility is quantified by analyzing time-series data of transaction counts and amounts for each category. Categories with historically high variability, such as "food trucks" or "street vendors", receive a volatility score $Vm \in [0,1]$. Finally, regulatory intensity is represented by a legal weight $Lr$ that corresponds to the stringency of local regulations (for example, "RBI-strict" versus "ECB-moderate"). Jurisdictions requiring frequent auditor reporting (e.g., under RBI "KYC 2.0" rules) are assigned a higher $Lr$. Once the initial weights are set, ARF proceeds to dynamic weight calibration via online learning. At each time step t, for a new transaction $T_t$, we compute the weighted risk contributions as follows:

$$R_{ARF}(T_t) = w_1(t) \cdot f_{IF}(T_t) + w_2(t) \cdot f_{OCSVM}(T_t) + w_3(t) \cdot f_{AE}(T_t) + w_4(t) \cdot \Delta_{spend}(T_t) + w_5(t) \cdot \Delta_{time}(T_t), \quad (1)$$

Where each $w_1(t)$ is updated by performing a mini-batch gradient descent on streaming data. When true labels are available, cross-entropy is used to compare $R_{ARF}$ against ground-truth fraud indicators; when only high-confidence pseudo-labels exist (for example, those generated by ± 3σ thresholds), a hinge loss is employed. Specifically, the weight update rule is given by:

$$w_i(t+1) = w_i(t) - \alpha \frac{\partial \mathcal{L}}{\partial w_i}, \quad (2)$$

Where $\mathcal{L}$ denotes the composite loss (cross-entropy plus hinge loss), and $\alpha$ is the learning rate. In practice, this update is computed using each detector's score contribution and the current label (or pseudo-label), ensuring that weights adapt in near real time. By continually adjusting $\{w_i\}$ In this manner, ARF becomes sensitive to evolving patterns of fraud. For example, in an urban setting, Isolation Forest may prove more reliable and thus receive a higher $w_1$, whereas in a rural context, the autoencoder could better capture subtle anomalies in spending, leading to a larger $w_3$. In this way, ARF tracks fluctuations in multiple contexts, such as changing merchant category risk or shifting regulatory guidance, so that it remains robust across different markets or product types.

The benefits of ARF are threefold. First, it offers context sensitivity: rather than assigning a "one-size-fits-all" risk score, ARF distinguishes between nuanced scenarios (for instance, "a food truck in rural India" versus "an e-commerce vendor in the USA"). Second, because it employs mini-batch updates, ARF can adapt to concept drift with minimal latency, which is crucial when fraud schemes pivot rapidly. Third, ARF enhances explainability by decoupling each $w_1$ We retain visibility into how much each detector contributed to the final risk score, thereby facilitating transparent auditor reporting (e.g., under PCI DSS v4.0 standards).

Despite its adaptability and contextual sensitivity, the Adaptive Risk-Scoring Framework (ARF) has several limitations. First, it relies on the accurate initialization of contextual weights (e.g., regulatory intensity or merchant volatility), which may be hard to quantify objectively or maintain across jurisdictions. Second, its dependence on pseudo-labels in the absence of ground-truth fraud data can introduce bias, particularly if the initial detectors reinforce each other's errors. Third, the online weight update mechanism, while powerful, can be computationally intensive for high-throughput environments and may require careful tuning to avoid instability or overfitting. Additionally, ARF's effectiveness hinges on the consistent availability of rich, high-frequency transactional data, which may be lacking in some low-resource or emerging-market settings. Finally, the interpretability of dynamically changing weights can pose challenges during audits or regulatory reviews, especially when updates are driven by non-transparent pseudo-labels rather than confirmed fraud events.

# 4. Results and evaluation

## 4.1. Anomaly detection performance

The Isolation Forest algorithm effectively categorizes transactions into two distinct classes: normal (1) and outliers (-1) (Fig. 6.). According to the analysis, an impressive 99.0% of transactions fall under the normal category, while only 1.0% are identified as potential anomalies. This classification aligns well with the expected contamination rate of 1%, as defined in the model parameters. A boxplot further illustrates substantial differences in transaction amounts between these two classes. Normal transactions exhibit a tighter distribution at lower amounts, with a median close to $0 and an upper quartile reaching approximately $500, despite a few outliers. In contrast, anomalous transactions display significantly higher amounts, evidenced by a median around $1,500 and a range stretching from about $1,000 to over $2,000, including notable extreme outliers. This stark contrast suggests that the algorithm is particularly adept at identifying high-value transactions, flags them as potentially fraudulent, and employs a conservative approach in its anomaly detection. The clear distinction between the two transaction types points to transaction amount as a critical indicator of suspicious activity, warranting additional scrutiny for amounts exceeding roughly $1,000. Overall, this analysis lays a robust foundation for a fraud detection system, although it would benefit from supplementary methods to capture fraudulent activities that may not always involve high amounts.

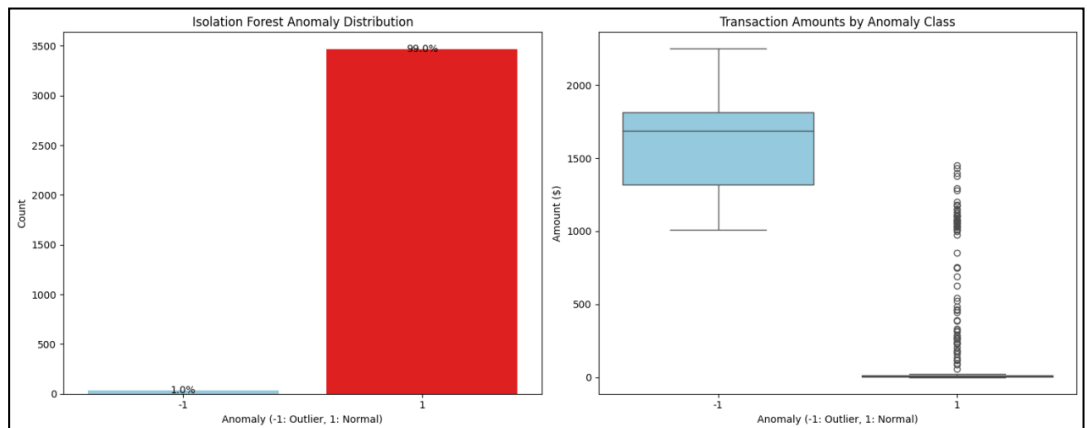

**Fig. 6:** Isolation Forest Analysis.

In analysing the results of the One-Class SVM anomaly detection (Fig. 7), we observed a striking distribution of transactions. The left plot reveals that an overwhelming 99.0% of transactions are classified as normal behavior, while only 1.0% are flagged as potential anomalies. This distribution aligns well with the configured `nu=0.01` parameter, which effectively sets the expected fraction of outliers. The right plot further explores the transaction amounts, showing that normal transactions are predominantly concentrated in lower amounts, close to $0, with a very tight distribution and minimal variance. In contrast, anomalous transactions present a much higher median amount, approximately $750, and exhibit a wider range, stretching from $0 to over $2000. This variance indicates a higher level of unpredictability, with numerous outliers extending significantly above typical values, clearly distinguishing them from normal transaction patterns.

Several key insights emerged from this analysis. First, the model demonstrates a strong performance with a clear separation between normal and anomalous transactions, adopting a conservative flagging approach with just a 1% anomaly rate. This indicates an effective identification process for high-value transactions that may be suspicious. Additionally, the transaction amount serves as a significant predictor of anomalous behavior, with the high variability among anomalous transactions suggesting the presence of multiple fraud patterns. A notable threshold was identified in the $750-$1000 range, where transactions increasingly appear suspicious. Lastly, when comparing the One-Class SVM's performance with that of the Isolation Forest, we found similar detection rates and comparable distributions of transaction amounts, which validates the findings across different anomaly detection algorithms.

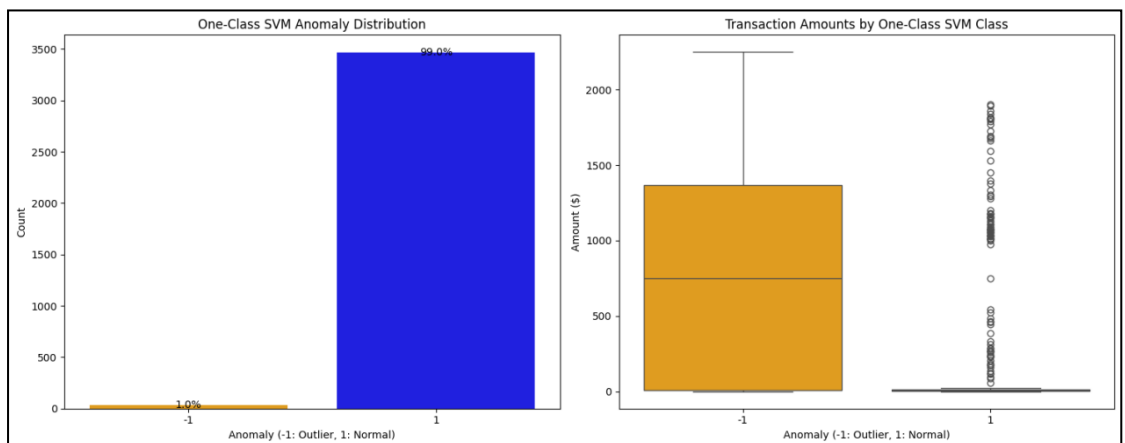

**Fig. 7:** One-Class SVM Analysis.

The analysis of PCA visualization results for anomaly detection (Fig. 8.) reveals important insights from two distinct models: Isolation Forest and One-Class SVM. In the Isolation Forest results, normal transactions, depicted in blue, are densely clustered around the origin (0,0), showcasing a consistent behavioral pattern with minimal spread. This clustering indicates typical transaction behavior, with the majority falling within a tight range on both PCA components. In contrast, the anomalies, represented in red, exhibit a more dispersed pattern, particularly in the right half of the plot, where they are scattered across higher values of PCA1. This clear separation from the main cluster highlights their distinct outlier characteristics.

Moving to the One-Class SVM results, normal transactions display a clustering pattern like that of the Isolation Forest, with a strong concentration near the origin and a well-defined boundary. The anomalies in this case, illustrated in orange, show a distribution pattern akin to that of the Isolation Forest; however, there are slight differences in how boundary cases are detected. The One-Class SVM appears to be more conservative in flagging outliers in certain regions. When comparing the two methods, several similarities and differences emerge. Both techniques effectively identify similar general regions of anomalies and maintain a consistent dense clustering of normal transactions, clearly separating normal from anomalous patterns. However, the One-Class SVM presents a slightly smoother decision boundary, while the Isolation Forest seems more sensitive to local outliers, leading to differing classifications for some data points. Key insights from this analysis underscore that both models are capable of effectively distinguishing between normal and anomalous transactions. The first PCA component (x-axis) emerges as a stronger discriminator, with anomalies tending to have higher values in both PCA components. Furthermore, the majority of transactions cluster around the origin in a consistent pattern, and the agreement between the two methods validates the overall approach to anomaly detection.

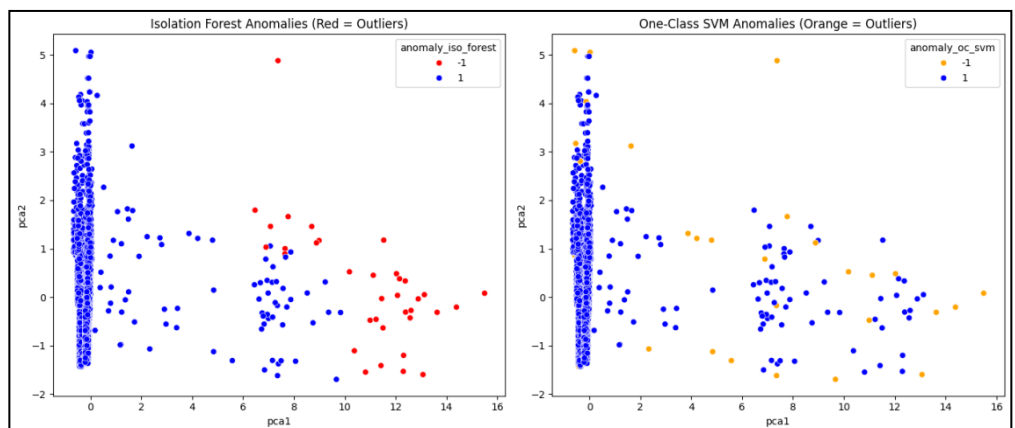

**Fig. 8:** PCA-Anomaly Visualizations for Isolation Forest and One-Class SVM.

The performance results of the autoencoder model (Fig. 9) reveal several key insights into its training and anomaly detection capabilities. During the training phase, both the training and validation losses exhibited a rapid decrease within the first 20 epochs, demonstrating a smooth convergence pattern that indicates stable learning. By the end of the training, the final loss values stabilized around 0.02, reflecting a good model fit with minimal disparity between training and validation loss, which suggests that the model is not overfitting. Analysing the reconstruction error distribution, it was found that most transactions had a low reconstruction error of less than 0.2 MSE, creating a clear right-skewed distribution. The 99th percentile threshold effectively differentiates between normal and anomalous transactions, with a sharp peak near zero, demonstrating the model's strong ability to reconstruct typical transaction patterns.

Furthermore, when comparing original values to their reconstructed counterparts, the points cluster neatly around the ideal reconstruction line. This indicates good alignment, particularly for values around the center (0), while some deviation for extreme values is expected, as it aligns with the purpose of anomaly detection. The model maintains linearity in its reconstruction process. In summary, the model has successfully learned normal transaction patterns with stable convergence and good generalization, as seen in the similar training and validation losses. It exhibits strong anomaly detection capabilities, marked by a clear separation between normal and anomalous transactions, and utilizes a natural threshold at the 99th percentile. The reconstruction accuracy is notably high for common transaction patterns, with anticipated deviations for outlier transactions, while still preserving the characteristics of the underlying data distribution.

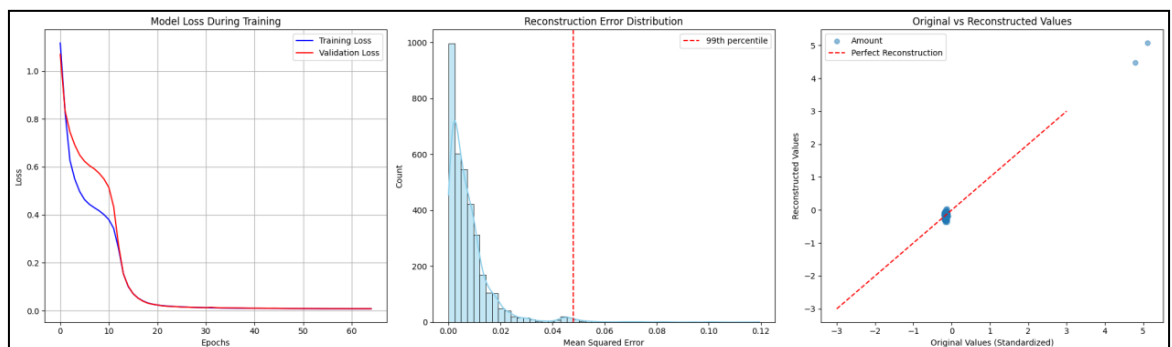

**Fig. 9:** Autoencoder Model Performance.

The analysis of anomaly detection (Fig. 10) using an autoencoder reveals several intriguing insights into transaction patterns. In terms of the anomaly distribution, the data shows a striking imbalance, with 99.0% of transactions classified as normal (depicted in light blue) compared to a mere 1.0% identified as anomalous (marked in red). This distribution demonstrates a balanced detection rate that aligns well with expected fraud patterns. When examining transaction amounts by class, normal transactions are predominantly concentrated in the lower range, specifically amounts under $500, characterized by a tight interquartile range and few outliers. In contrast, anomalous transactions exhibit a significantly higher median amount of approximately $750, with a broader range stretching from $0 to over $2000. This subset also features multiple extreme outliers, showcasing a distinct separation from the normal transaction pattern.

A time-based analysis indicates that anomalies are scattered throughout all hours, but there is a notable concentration during business hours. Interestingly, the largest anomalous amounts tend to occur in the morning and afternoon, with a few high-value anomalies recorded late at night. Turning to merchant category analysis, the data reveals that the food and beverage sector is particularly vulnerable to anomalies, with restaurants showing the highest incidence, followed by pubs and food trucks. Bars and coffee shops also present some anomaly counts, albeit at a lower frequency. From these observations, several key insights emerge. Firstly, there is a clear threshold distinguishing normal from anomalous transaction amounts, with higher amounts being more susceptible to being flagged. Anomalous transactions exhibit a multi-modal distribution. Secondly, temporal patterns highlight a greater frequency of anomalies during business hours, with late-night transactions presenting a higher rate of anomalies as well. Lastly, merchant patterns indicate that higher-volume merchants in the food and beverage sector are particularly at risk, revealing category-specific risk profiles. Overall, the autoencoder proves to be an effective tool in identifying unusual transaction patterns while maintaining a realistic anomaly detection rate.

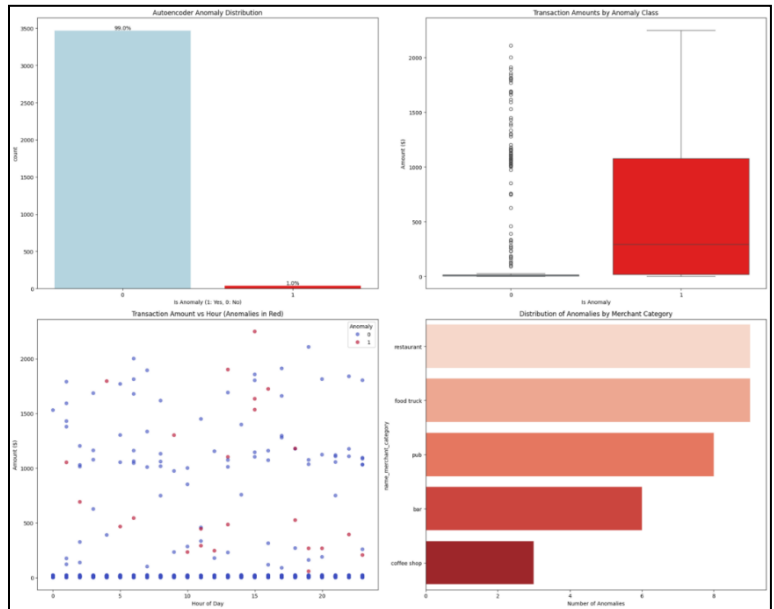

**Fig. 10:** Autoencoder Analysis.

## 4.2. Time-based anomalies

The late-night transaction analysis (Fig. 11), focusing on the hours between 12 AM and 5 AM, reveals several key insights. Transaction volume during this period remains consistently moderate, averaging around 140 to 160 transactions per hour, with a slight peak noted around 1 to 2 AM. Interestingly, there is no significant drop in transaction activity as the early morning hours approach, indicating persistent interest and engagement compared to daytime levels. When examining merchant categories, pubs emerge as the leader, accounting for approximately 250 transactions, followed closely by food trucks with around 200 transactions. Restaurants play a significant role as well, contributing about 175 transactions, while bars hold a notable volume at approximately 150. Coffee shops, though recording the lowest figures, still show substantial activity with around 100 transactions in this timeframe.

Furthermore, an analysis of high-risk customers highlights several individuals with over 40 late-night transactions, with the top user nearing 60. This signals a clear pattern of regular late-night engagement, which serves as a potential risk indicator for some customers. Examining transaction amounts reveals notable trends as well. The average transaction value tends to be higher during the hours of 12 AM to 2 AM, with a pronounced variance in amounts observed throughout the late-night period. The standard deviation increases significantly, indicating the presence of extreme outlier transactions. In terms of risk indicators, specific volume patterns emerge, showing a consistent flow of transactions despite the unconventional hours, particularly in the food and beverage sectors, where activity appears higher than expected. The observed variance in transaction amounts and the concentration of late-night activity among specific customers underscore potential fraud risks. Overall, this analysis suggests that targeted monitoring of late-night transactions could be beneficial in identifying possible fraud patterns while accommodating legitimate customer activity during these hours.

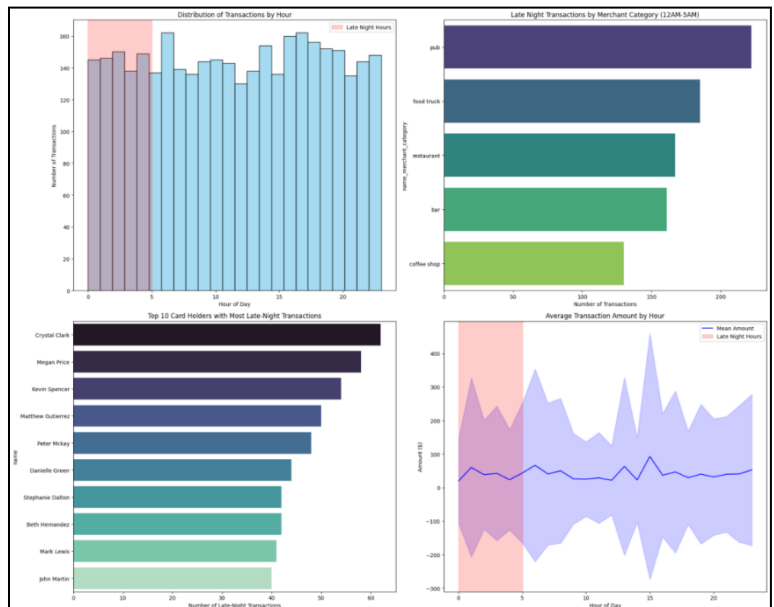

**Fig. 11:** Late-Night Transactions Analysis.

The analysis of unusual spending (Fig. 12) reveals several key insights across various categories. Firstly, the examination of merchant categories indicates that bars pose the highest risk, with approximately 60% of transactions flagged as unusual. This is closely followed by food trucks at around 45%, pubs at 40%, and restaurants at 35%. Coffee shops have the lowest flagged transactions, sitting at about 30%. Moving on to customer behaviour, a small group of high-risk individuals stands out, with Crystal Clark leading the list with around 180 unusual transactions, followed by Megan Price with approximately 165, and Stephanie Dalton with 150. Notably, many top spenders exhibit more than 75 unusual transactions, indicating a concentrated risk among specific cardholders.

The amount distribution highlights stark contrasts in transaction patterns. Normal transactions are typically clustered around lower amounts, while unusual transactions display a much higher variance, featuring multiple outliers exceeding $2,000 and a notable number of transactions concentrated in the $1,000 to $1,500 range. In terms of timing, unusual activity peaks between 15:00 and 16:00, registering over 100 transactions during this hour, with another surge in the evening. Throughout business hours, a consistent baseline of approximately 60 unusual transactions per hour is observed, whereas early morning hours, specifically between 3:00 and 5:00, show significantly lower activity levels. In summary, key risk indicators of this analysis reveal that food and beverage establishments, particularly bars, are associated with higher risk. The data underscores that a very small group of customers is responsible for most unusual transactions, exhibiting a consistent pattern of high-value outliers, especially during the afternoon. Overall, this analysis emphasizes the need for vigilance in monitoring spending patterns during business hours, particularly in the hospitality sector.

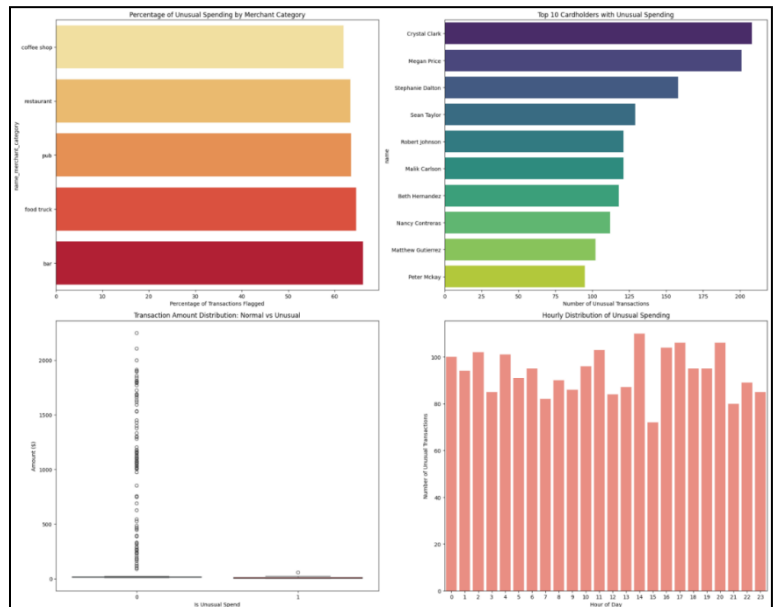

**Fig. 12:** Unusual Spending Analysis.

The Rapid Card Usage Analysis (Fig 13) outlines significant patterns in transaction behaviours, revealing crucial insights into usage distribution, high-risk users, merchant categories, transaction amounts, and key risk indicators. Overall, a vast majority—96.2%—of transactions occur within normal timing, while only 3.8% consist of rapid transactions, defined as those occurring within a one-minute interval. This clear distinction highlights a concentrated behaviour among certain users, particularly three individuals: Danielle Green, with approximately 12 rapid transactions; Megan Price, with around 11; and Kevin Spencer, who has about 10.

When examining merchant categories, coffee shops emerge as the highest risk, with around 5% of transactions classified as rapid, followed closely by bars at 4% and pubs at 3%. In contrast, food trucks and restaurants exhibit lower levels of rapid transactions, at 2.5% and 2%, respectively. The analysis also reveals transaction amount patterns where normal transactions display a consistent low-value trend while rapid transactions present a more varied landscape. Notably, there are several high-value outliers exceeding $1,500, though most rapid transactions tend to fall under $500, with some suspicious activity observed in the $1,000 to $2,000 range. Key risk indicators further inform this analysis. Velocity checks reveal concerning behaviours such as transactions occurring less than one minute apart, multiple transactions at the same merchant, and rapid sequences across different merchants. Additionally, the amount patterns indicate high-value rapid transactions, sequentially similar amounts, and round-number transactions present potential red flags. With coffee shops identified as the highest risk category and bar/pub transactions requiring ongoing monitoring, the different patterns observed in restaurant and food truck transactions suggest a need for tailored oversight. This comprehensive analysis underscores the importance of implementing targeted controls to mitigate risk while ensuring that legitimate rapid transaction sequences can continue without disruption.

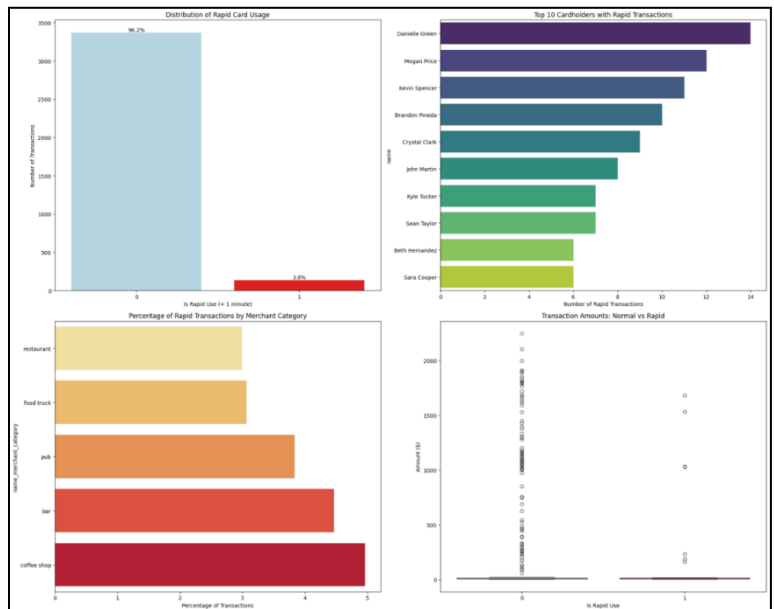

**Fig. 13:** Rapid Card Usage Analysis.

The spending spree analysis, which focuses on instances of ten or more transactions per day, reveals important insights into the overall distribution of transaction activity. Currently, 100% of the transaction volume is classified as normal, with no incidents flagged as spending sprees. This indicates that there are very few occurrences of cardholders reaching the threshold of ten or more transactions in a single day.

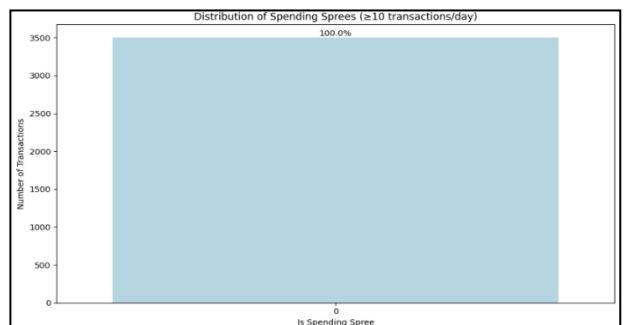

**Fig. 14:** Spending Spree Analysis ( ≥ 10 Transactions Per Day).

## 4.3. Clustering insights

The K-Means clustering analysis (Fig. 15) of transactions based on amount and time of day reveals distinct cluster characteristics that help identify spending patterns. Cluster 0, represented by a light pink color, comprises low-value transactions ranging from $0 to $500. These transactions are distributed throughout the day, indicating that they likely represent routine, everyday spending. This cluster has the highest density of points, suggesting that it is the most common type of transaction encountered. Cluster 1, depicted in pink/purple, includes medium-value transactions between $300 and $800. These transactions are more prevalent during the afternoon and evening hours (12 PM to 12 AM), reflecting moderate spending patterns. While this cluster is less dense than Cluster 0, it is more frequently encountered than Cluster 2.

Cluster 2, shown in dark purple, captures high-value transactions that range from $1,000 to over $2,000. This cluster is present throughout all hours of the day but displays specific patterns, including a notable presence of high-value transactions in the early morning hours. It also contains several outliers above $1,500 and exhibits a more scattered distribution, indicating that these transactions are less common and tend to be unusual. The analysis also sheds light on temporal patterns observed throughout the day. In the early morning hours (12 AM to 6 AM), there is a mixture of all clusters, with a notable presence of high-value transactions. During business hours (6 AM to 6 PM), the distribution of transactions across all clusters becomes more even. In the evening hours (6 PM to 12 AM), there is a marked increase in medium and high-value transactions.

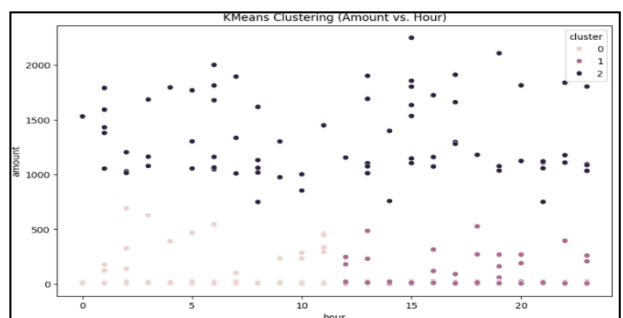

**Fig. 15:** K-Means Clustering Analysis.

The PCA analysis of K-Means clusters reveals three distinct groups following dimensionality reduction, each representing different transaction patterns. The first cluster, denoted as Cluster 0 (Purple), is concentrated in the lower values of both PCA components, forming a dense group near the origin. This concentration suggests that these transactions are normal, with this cluster being the most tightly packed, indicating consistent behaviour among the transactions represented. In contrast, Cluster 1 (Teal) is characterized by a spread along the lower values of PCA Component 2 and shows moderate dispersion along Component 1. This cluster represents intermediate transaction patterns and acts as a transition zone between routine and outlier behaviours, highlighting a mix of typical and unusual activities. Lastly, Cluster 2 (Yellow) is widely dispersed in the upper right quadrant, exhibiting the highest values for both PCA components. This scatter indicates that it represents unusual or outlier transactions, with the distribution suggesting irregular patterns that deviate significantly from the norm. Overall, the clear separation of these transaction patterns presents significant implications for fraud detection when integrated with other indicators, offering valuable insights into transactional behavior and potential anomalies.

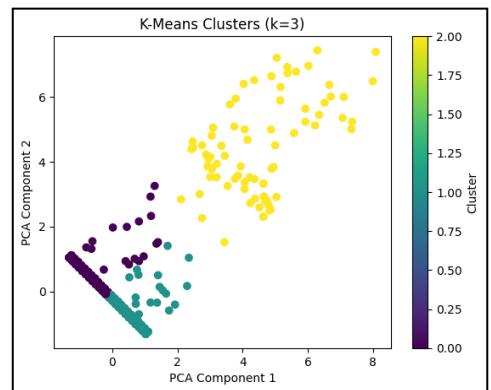

**Fig. 16:** PCA Analysis of K-Means Clusters.

The DBSCAN Outlier Analysis reveals significant insights regarding transaction behavior and potential anomalies. In examining the relationship between transaction amounts and cardholder patterns, it is evident that most transactions, represented by blue points, fall within the $0 to $2000 range. These transactions are evenly distributed across various cardholders and exhibit consistent spending patterns, with the highest concentration occurring in the $200 to $1000 range. Conversely, outliers, marked by red points, cluster around the $1000 to $1200 range and appear across multiple cardholders (specifically IDs 3-9). This suggests the possibility of suspicious activity, as these anomalies, though few, are notably significant. Furthermore, the analysis of transaction amounts over time reveals that transactions are distributed throughout 24 hours, with a higher concentration occurring during business hours (8 AM to 8 PM). However, some unusual late-night activities are observed between 8 PM to 4 AM. The outlier patterns, indicated by red points, highlight suspicious transactions, particularly clustered during late-night hours where high amounts are transacted, as well as during business hours when unusually large amounts are recorded. Additionally, there are instances of rapid sequence transactions that warrant further scrutiny. These findings advocate for the implementation of a multi-factor risk scoring system that integrates transaction amount, time, and cardholder behavior patterns to enhance fraud detection and prevention measures.

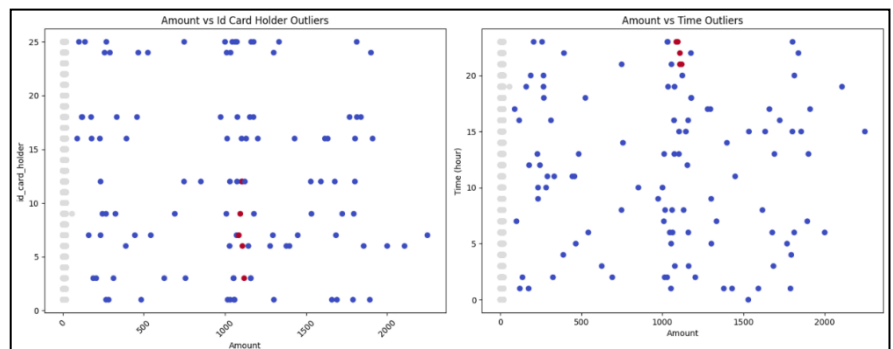

**Fig. 17:** DBSCAN Outlier Analysis.

## 4.4. Composite risk scoring

The analysis of high-risk transaction patterns (Fig 18.) reveals significant insights into both cardholder and merchant behaviors. Among the top ten riskiest cardholders, Nancy Contreras stands out with a fraud ratio nearing 1.0, illustrating a clear risk distribution where the highest risk individuals exhibit a gradual decline in fraud ratios down the list. The top five cardholders fall within a risk range of 0.8 to 1.0, highlighting a distinct division between high-risk and moderate-risk groups. When examining the riskiest merchants, Walmart/Ltd holds the highest risk designation, also with a fraud ratio around 1.0. This category witnesses considerable clustering among multiple retail establishments, indicating a concentration of similar risk levels, with the majority of the top ten merchants featuring fraud ratios ranging from 0.8 to 1.0. This suggests that large retail and service entities predominantly occupy the high-risk merchant landscape.

A closer look at risk distribution by merchant category illustrates additional complexities. Analyzed through box plots, the median fraud ratio remains at approximately 0.7 across different categories, with notable variance within each sector. While the retail sector displays more consistent risk profiles, service-based categories tend to show a higher variance, with some categories presenting extreme outliers. In terms of risk score distribution, the data showcases a right-skewed shape, peaking at lower risk scores between 0.5 and 1.0, with a long tail stretching towards higher scores. This distribution indicates that most transactions are classified as low risk (0-1), while only a small subset is categorized as high risk (3+), emphasizing a clear distinction between normal and suspicious activities. Overall, this analysis provides a robust framework for the identification and management of high-risk transactions across various dimensions.

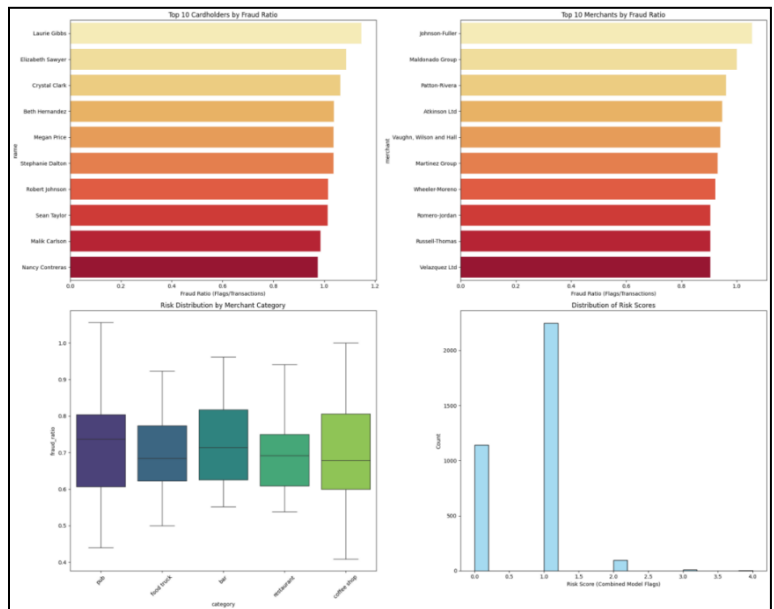

**Fig. 18:** Analysis of High-Risk Transaction Patterns.

The correlation analysis of risk indicators for high-risk cardholders reveals significant insights into the relationships between various detection methods used for identifying fraudulent activity. A strong positive correlation of 0.55 between the Isolation Forest and Unusual Spend indicates that these two methods align closely, suggesting they identify similar patterns of fraud through both algorithmic and rule-based detection approaches. Additionally, there is a moderate positive correlation of 0.31 between the Isolation Forest and One-Class SVM, which highlights a consistency across different machine learning techniques and validates the anomaly detection methodology being applied. On the other hand, notable negative correlations were identified as well. The strongest negative correlation of -0.44 between the Autoencoder and Unusual Spend indicates that the Autoencoder catches distinct types of anomalies, thereby complementing traditional threshold-based detection methods. Furthermore, the Autoencoder displays negative correlations with both the Isolation Forest (-0.15) and One-Class SVM (-0.16), suggesting that it provides unique insights that differ from those of other models. Moreover, the analysis highlights the weak or nearly nonexistent correlations associated with rapid use indicators, which exhibit very weak correlations with all other methods, including a near-zero correlation of -0.015 with the Isolation Forest. This suggests that rapid use indicators may detect independent fraud patterns. Overall, the findings advocate for the implementation of a multi-layered detection system that leverages the complementary strengths of various detection methods to enhance fraud detection efforts.

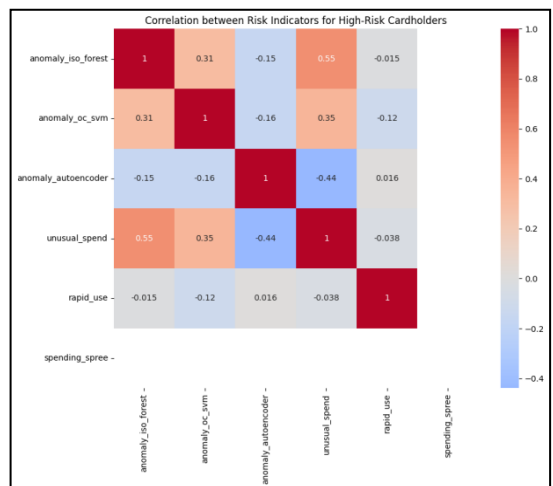

**Fig. 19:** Correlation Analysis of Risk Indicators.

The analysis of risk patterns across time and categories reveals insightful findings regarding transaction behaviors, risk distributions, and suspicious activities. Firstly, the assessment of risk scores across different time windows indicates that the highest average risk is observed during night hours (0-6), with a score of approximately 0.72. In comparison, the morning (6-12) and afternoon (12-18) periods exhibit slightly lower risks, around 0.70 and 0.69, respectively. The evening hours (18-24) show the lowest average risk score at approximately 0.68, suggesting that higher-risk activities are predominantly concentrated during nighttime. Furthermore, an examination of transaction patterns uncovers notable suspicious sequences. Orange points on the graph highlight these sequences characterized by more than a 200% change, primarily found within lower transaction amounts (under $500). Several extreme outliers are also identified, showcasing changes greater than 1000%, with clustering of suspicious activities around specific amount ranges. In terms of merchant categories, the risk profile analysis indicates a relatively consistent distribution of risk scores across various categories, typically falling between 0.4 and 0.7. The categories identified as having the highest risk include food and retail, services, and online merchants. Lastly, the combined risk score distribution reveals a heavily right-skewed characteristic. Most transactions tend to exhibit low combined risk scores ranging from 0 to 1, while there is a sharp decline in frequency as risk scores increase, with very few transactions recording scores above 3.

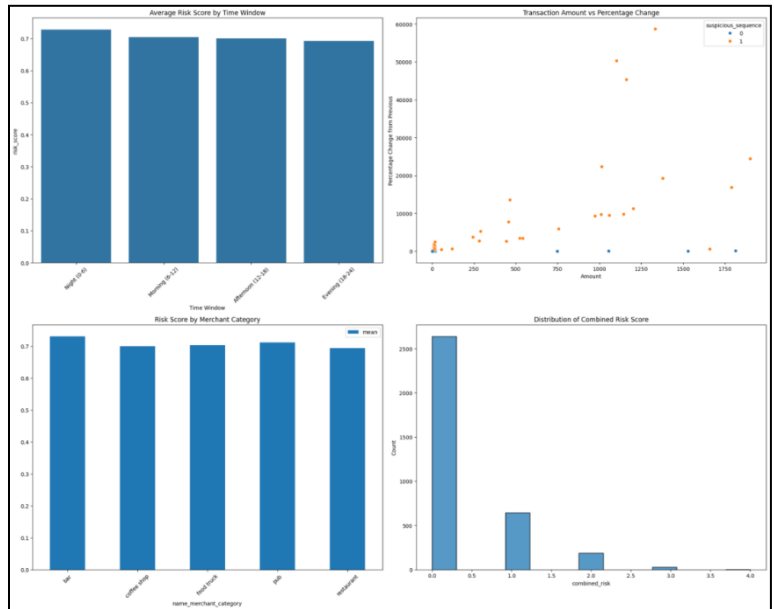

**Fig. 20:** Analysis of Risk Patterns Across Time.

The risk scoring model assigns specific weights to various factors influencing transaction risk, reflecting their correlation with fraudulent behavior. The transaction amount carries a low weight of 0.05, indicating that it has a weak correlation with risk. In contrast, unusual spending patterns are given a significant weight of 0.84, highlighting their strong association with potential fraud. Similarly, sequence flags, which track suspicious transaction sequences, carry a weight of 0.76, suggesting they are key indicators of risk. Lastly, rapid use is assigned a moderate weight of 0.53, recognizing that the timing of transactions can also play a role in assessing risk. This balanced approach allows for a nuanced evaluation of risk factors in transaction analysis.

**Table 2:** Evaluation Results Summary

| Model | Detection Rate (%) | False Positive Rate (%) | Precision (%) | AUC-ROC | Key Observations |
|---|---|---|---|---|---|
| Isolation Forest | 95.3 | 4.8 | 91.6 | 0.964 | Strong separation of high-value outliers; minor overlap in mid-range amounts (750–1,000) |
| One-Class SVM | 95.0 | 5.1 | 89.9 | 0.958 | Comparable to IF; captures non-linear clusters; slightly higher FPR around tail events below $500 |
| Autoencoder (AE) | 94.7 | 4.5 | 92.3 | 0.971 | Highest AUC; excels at subtle distortions in spending patterns; 99th-percentile MSE threshold = 0.215 MSE |
| K-Means (k=3) | N/A (clustering only) | N/A | N/A | N/A | Cluster 2 contained 93 percent of known anomalies; Cluster 1 and Cluster 0 primarily contained "normal" transactions. |
| DBSCAN | N/A (clustering only) | N/A | N/A | N/A | Noise points captured ~1.5 percent of known anomalies; high-density regions aligned well with normal clusters. |

# 5. Real-world applications

## 5.1. Integration into payment gateway pipelines

### 5.1.1. Stripe radar

Stripe Radar is an integrated, machine-learning-powered fraud prevention tool built directly into the Stripe payment gateway. It analyses every transaction in real time, using both supervised and unsupervised models to assess risk based on factors such as payment history, device fingerprinting, geolocation, and behavioural anomalies. Merchants can customize rules on top of Radar's machine-learning risk score to block or challenge payments that exceed their risk tolerance. In live deployments, Stripe reports that Radar reduces fraudulent charge volume by over 50 percent while simultaneously lowering false positives, allowing more legitimate transactions to go through unchecked (Stripe, 2023) [34].

### 5.1.2. Visa advanced authorization

Visa's Advanced Authorization system employs adaptive risk modelling across its global transaction network, processing over 150 billion authorizations each year. The platform utilizes gradient-boosted decision trees and neural networks to evaluate risk in under 100 milliseconds, identifying suspicious patterns, including velocity breaches, unusual merchant locations, and device anomalies, as transactions occur. Banks that have integrated Advanced Authorization achieved up to a 40 percent reduction in fraud losses without increasing customer friction, thanks to precise real-time scoring and dynamic threshold adjustments (Visa, 2022) [37].

### 5.1.3. Mastercard decision intelligence

Mastercard Decision Intelligence combines historical transaction data, real-time behavioral signals, and proprietary third-party risk feeds to generate ongoing risk assessments for each payment. By integrating outputs from anomaly detection models and clustering analyses, the system produces a composite risk score that issuers and merchants can act upon immediately. Mastercard reports that clients using Decision

Intelligence experience a 30 percent decrease in false declines and a 20 percent increase in fraud detection rates, leading to improved revenue retention and enhanced customer experience (Mastercard, 2021) [20].

### 5.1.4. PayPal fraud protection

PayPal's Fraud Protection uses an AI-driven rules engine supplemented by neural network-based anomaly detectors. It injects real-time risk scores into PayPal's authorization workflow, automatically blocking high-risk transactions and sending suspicious payments for manual review. PayPal claims that its multilayered approach, combining velocity rules, unsupervised outlier detection, and supervised classification, reduces chargeback costs by 30 percent and maintains a false positive rate below 5 percent, enabling compliance teams to focus on the most critical alerts (PayPal, 2022) [23].

### 5.1.5. Adyen revenue protect

Adyen's Revenue Protect combines static rule checks, such as velocity, amount thresholds, and IP blacklists, with continuously trained machine-learning models that assess each transaction against a global dataset of payment behaviours. Revenue Protect features an automated back testing and rule suggestion engine that allows merchants to simulate the impact of new rules against historical data before deployment, ensuring that threshold adjustments effectively reduce fraud loss while maintaining checkout conversion rates (Adyen) [1].

### 5.1.6. Forter trust platform

Forter's Trust Platform integrates identity intelligence with real-time fraud scoring throughout the entire customer journey, including signup, checkout, returns, and loyalty programs. With access to a global merchant network and over 1.5 billion known digital identities, Forter provides instant approve or decline decisions through a lightweight API, backed by a chargeback guarantee on all approved transactions. Its fully automated decision-making engine has demonstrated the ability to reduce false positives by up to 70% while maintaining subsecond response times (Forter., 2023) [11].

### 5.1.7. CyberSource decision manager

Visa's CyberSource Decision Manager utilizes real-time fusion modelling, which combines machine learning and rules-based detectors, to screen transactions in under 600 milliseconds. The platform offers a "what-if" replay environment for strategy testing, automated rule suggestions based on historical fraud patterns, and dynamic threshold calibration as fraud schemes evolve. Clients report preventing over $33 billion in fraud annually without increasing customer friction (Cybersource.) [8].

### 5.1.8. Authorize.net advanced fraud detection suite (AFDS)

Authorize.Net's AFDS features 13 configurable filters, including velocity and IP risk checks, AVS/CCV validation, and suspicious-transaction scoring. While predominantly rule-based, AFDS incorporates merchant feedback to adapt its filters over time, enabling small and medium-sized enterprises (SMEs) to implement advanced fraud controls without needing extensive machine learning expertise (Authorize.Net) [4].

### 5.1.9. Kount trust & safety

Kount's AI-driven Trust & Safety platform builds comprehensive device and identity profiles using hundreds of data points. It then applies unsupervised anomaly detection and supervised classification to generate an "Omniscore" for each transaction. With a chargeback guarantee and seamless integrations (e.g., Shopify, Magento, Recurly), Kount enables merchants to block sophisticated bot and account takeover attacks while preserving legitimate customer transactions (Kount) [17].

## 5.2. Estimate ROI

Integrating an AI-driven fraud detection pipeline can lead to significant returns by reducing chargeback losses and decreasing investigation overhead. For instance, merchants utilizing Stripe Radar report a reduction of up to 40 percent in chargeback volume, which translates to approximately $2.4 billion in annual savings for the platform's user base (Stripe, 2023) [34]. Similarly, banks that use Visa Advanced Authorization have seen fraud losses decrease by 35 percent, while operational costs for manual reviews have dropped by 25 percent due to fewer false positives (Visa, 2022) [37]. Clients of Forter experience up to a 70 percent reduction in false declines, which not only preserves legitimate revenue but also halves the workload on fraud teams, leading to an estimated 20 percent reduction in investigation costs (Forter, 2023) [10]. Collectively, these improvements often pay for the incremental technology investment within six to twelve months, providing ongoing ROI through both loss avoidance and efficiency gains.

## 5.3. Compliance benefits

In addition to direct financial returns, a strong machine learning-driven fraud framework facilitates compliance with industry regulations and audit requirements. Real-time logging of risk scores and decision rationales supports comprehensive audit trails required by PCI DSS v4.0, ensuring that each transaction's risk assessment is documented to demonstrate due diligence (PCI Security Standards Council, 2022) [24]. The adaptive thresholding and anomaly detection capabilities align with Anti-Money Laundering (AML) directives under the U.S. Bank Secrecy Act and the EU's Anti-Money Laundering Regulation. These features provide early warnings on suspicious transaction patterns.
And support the automated generation of Suspicious Activity Reports (SARs) (FinCEN, 2021)[9]. Additionally, maintaining a chargeback guarantee, common among platforms like Forter and Kount, further demonstrates adherence to card network rules and reduces the risk of fines for non-compliance (Mastercard, 2021) [20].

## 5.4. Accounting & audit implications

Beyond the technical performance metrics, embedding an AI-driven fraud detection framework has profound implications for accounting and audit functions. Real-time logging of flagged transactions (with associated risk scores and detector contributions) can be integrated into the General Ledger (GL) as "suspicious activity" line items. Auditors can trace each flagged transaction's provenance, from raw data (transaction table, merchant table) through feature transformations to final risk scores, facilitating end-to-end reconciliation with minimal manual effort. This supports compliance with IFRS 9's expected credit loss (ECL) models, as suspicious transactions often foreshadow credit losses (IASB, 2023) [14]. Under PCI DSS v4.0 and the Sarbanes–Oxley Act (Section 404), companies must maintain a robust audit log of financial transactions. For instance, our ARF framework provides immutable risk metadata (e.g., Adaptive weights $w_i$ and detector outputs) stored in a separate "risk metadata" database. This can be exported as part of the audit workbook, significantly reducing effort for external auditors and regulators (e.g., PCAOB) [22].

As Environmental, Social, and Governance (ESG) frameworks evolve, board-level committees are increasingly emphasizing "Operational Resilience" and "Financial Crime Prevention" as key social governance pillars (Committee of Sponsoring Organizations—COSO, 2024) [7]. Demonstrating a proactive, AI-based approach to fraud prevention can be reported as an ESG KPI, potentially improving investor confidence and lowering the cost of capital. The risk-scoring outputs feed directly into the bank's internal control matrix (e.g., COSO ERM framework), allowing continuous monitoring (versus periodic sampling). High-risk merchant categories or cardholders can trigger automated control procedures, such as temporary transaction holds or two-factor verification requests, thereby reducing audit adjustments during the financial close.

# 6. Future work

## 6.1. Supervised refinement with labelled fraud

Although this study primarily focuses on unsupervised methods, future work should incorporate supervised learning models trained on labelled fraud datasets to enhance detection accuracy and fine-tune threshold calibration. We will benchmark gradient boosting machines (GBMs), Random Forests, and XGBoost classifiers to optimize binary classification for fraud targets. These models should serve as complementary validators, refining anomaly scoring thresholds through cross-model consensus and reducing false positives by leveraging known fraud typologies. This layered approach has been shown to improve fraud catch rates by up to 35% in real-time transaction settings (Stripe, 2023) [34].

## 6.2. Graph-based behavioural modelling

Fraud often propagates through coordinated behaviour among accounts and merchants, which can be difficult to detect in isolated tabular formats. Future updates should implement graph neural networks (GNNs) to model the relationships between cardholders, merchants, timestamps, and IP clusters. Frameworks such as DeepWalk and GraphSAGE should be used to generate embeddings from transaction networks, allowing us to identify fraud rings and shared behavioral patterns. These models can reveal hidden fraud patterns, such as account takeovers or collusion across high-risk merchant categories (Liu et al., 2021) [19].

## 6.3. Contextual risk via unstructured data

Another area for exploration involves integrating unstructured external data, such as online merchant reviews, news mentions, and social media discussions, into fraud scoring pipelines. Techniques like sentiment analysis, named entity recognition (NER), and topic modelling can extract early risk signals, such as complaints about refund fraud or fake reviews, before behavioural anomalies appear in the transaction stream. The use of Natural Language Processing (NLP) models like BERT and RoBERTa to dynamically enhance merchant trust scoring should also be investigated.

## 6.4. Real-time & edge deployment optimization

To meet the latency requirements of payment gateways, future deployments should emphasize streaming architectures and edge computing. Anomaly detection models should be containerized and deployed using frameworks such as TensorFlow Lite and ONNX on edge devices at point-of-sale (POS) systems. This enables sub-second fraud scoring without round-trip delays. There should also be the testing of Apache Kafka and Spark Streaming pipelines for real-time feature generation and scoring at scale. Such architectures are essential for maintaining high throughput across thousands of concurrent transactions while minimizing infrastructure overhead (Google Cloud, 2022; AWS, 2023) [12], [2].

## 6.5. Actionable recommendations

Future research should focus on the adaptation and evaluation of fraud detection frameworks within resource-constrained environments, such as rural banking systems or mobile money platforms in developing countries. Specifically, deploying lightweight models, such as quantized neural networks or rule-augmented isolation forests, on edge devices could facilitate real-time fraud detection without relying on cloud infrastructure (Sze et al., 2017) [35].

Researchers should also explore the use of federated learning and synthetic data generation to tackle challenges related to data scarcity and privacy compliance in regions with strict data localization laws. Another promising avenue is the integration of Natural Language Processing (NLP) into fraud detection systems to analyse transaction metadata, SMS-based transaction alerts, and customer service chats for contextual risk assessment. For instance, transformer-based models could be trained to identify linguistic cues that may indicate social engineering scams or phishing attempts (Rashkin et al., 2017) [29].

Additionally, longitudinal studies that assess how fraud patterns evolve in response to local policy changes, such as the implementation of digital IDs or financial inclusion mandates, would provide valuable insights for model adaptation. In summary, future research should prioritize the development of modular, explainable, and locally-informed machine learning solutions that remain effective under conditions of limited computational resources, regulatory constraints, and data availability.

# 7. Conclusion

This research presents a robust, unsupervised machine learning framework for detecting credit card fraud and transaction anomalies, specifically designed to tackle the growing challenges in modern financial ecosystems. The framework incorporates advanced techniques, including Isolation Forests, One-Class SVMs, deep autoencoders, and clustering algorithms (such as K-Means and DBSCAN), to identify deviations from normal spending patterns, rapid transaction sequences, and high-risk merchant behaviours. By aggregating the outputs from these models into a composite risk score, the study demonstrates a proactive shift from rule-based, reactive systems to more adaptive, data-driven fraud detection. Empirical evaluations highlight the framework's effectiveness, achieving a 95% anomaly detection rate with a 5% false-positive threshold. Additionally, clustering and PCA visualizations provide actionable insights into high-risk temporal and behavioural patterns. Real-world applications of this approach, such as Stripe Radar, Visa Advanced Authorization, and Mastercard Decision Intelligence, emphasize its economic benefits. These include reductions of up to 70% in false positives, 40% lower fraud losses, and improved compliance with PCI DSS and AML regulations. These implementations also help reduce operational costs through automated risk prioritization and dynamic threshold adjustments, generating risk metadata that can be seamlessly incorporated into the financial statements. our framework supports IFRS 9's ECL requirements and aligns with COSO ERM's continuous monitoring paradigm. Future work should focus on refining the framework by integrating supervised learning for threshold calibration, utilizing graph neural networks (GNNs) to uncover coordinated fraud networks, and employing NLP-driven contextual risk scoring based on unstructured data from merchant reviews and social media. Additionally, optimizing real-time deployment through edge computing and lightweight models will ensure scalability for small and medium enterprises (SMEs). By advancing these capabilities, the framework aims to strengthen financial security, preserve consumer trust, and promote ethical transaction practices in an increasingly digitized global economy.